\documentclass[runningheads]{llncs}

\usepackage{eccv}

\usepackage{eccvabbrv}

\usepackage{graphicx}
\usepackage{booktabs}

\usepackage[accsupp]{axessibility}  

\usepackage{amsmath,amsfonts,amssymb}
\usepackage{threeparttable} 
\usepackage{multirow}
\usepackage{subcaption}

\usepackage{hyperref}

\usepackage{orcidlink}

\begin{document}

\title{Incremental Online Scene Reconstruction by 3D Gaussian Triangulation} 


\author{Yanjin Zhu\textsuperscript{*}\inst{1}\orcidlink{0009-0006-4860-0269} \and
Shaofan Liu\textsuperscript{*}\inst{2}\orcidlink{0000-0002-2705-7117} \and
Jianke Zhu\textsuperscript{\dag}\inst{1,3}\orcidlink{0000-0003-1831-0106}} 

\authorrunning{Y.~Zhu et al.}

\institute{
  Zhejiang University, Hangzhou, China \and
  Hefei University of Technology, Hefei, China \and
  Shenzhen Loop Area Institute, Shenzhen, China \\
}

\maketitle
\renewcommand{\thefootnote}{*}
\footnotetext[1]{Equal contribution.}
\renewcommand{\thefootnote}{\dag}
\footnotetext[2]{Corresponding author.}

\begin{abstract}
Incremental scene reconstruction is essential for real-world applications. Although 3D Gaussian Splatting shows strong potential, most existing approaches require offline conversion of the optimized Gaussians into an intermediate implicit field for explicit mesh extraction, which hinders seamless integration with downstream tasks. To address this limitation, we propose a novel online framework that incrementally reconstructs and updates high-fidelity explicit meshes by directly triangulating a dense geometric Gaussian representation, which supports both high-quality rendering and incremental surface reconstruction. Moreover, we present a direct meshing algorithm that efficiently extracts and updates the mesh from the Gaussian set. To ensure mesh accuracy, we enforce a plane-based pulling constraint that dynamically aligns 3D Gaussian primitives to the approximated local surface. Furthermore, our framework significantly reduces memory and computational overhead during long-sequence processing by dynamically freezing fully optimized historical regions. Experiments on public datasets demonstrate that our method outperforms conventional Gaussian-based methods on both rendering quality and reconstruction accuracy.
  \keywords{Gaussian Triangulation \and Incremental Reconstruction \and Gaussian Splatting}
\end{abstract}

\section{Introduction}
\label{sec:intro}
3D scene reconstruction is a fundamental task with extensive applications in computer vision and robotics, notably Augmented Reality~\cite{schops2017real} and scene perception~\cite{hane20173d}. Typically, 3D scene reconstruction techniques can be categorized into two groups. One is implicit methods~\cite{mildenhall2021nerf,zhang2024neural}, and the other is explicit approaches~\cite{kerbl20233d,yu2024gaussian,guedon2024sugar}. Unfortunately, most high-fidelity reconstruction methods require processing the entire scene before generating the recovered meshes, which poses a significant bottleneck in real-world applications. For example, the memory constraints of computing devices make it infeasible to represent continuously expanding scenes. For time-critical tasks in autonomous robotics, the delay in obtaining a complete model is impractical, as early access to partial results is crucial for making immediate decisions. Consequently, incremental 3D scene reconstruction is urgently needed to overcome these fundamental limitations.

Traditional online scene reconstruction methods, such as KinectFusion~\cite{newcombe2011kinectfusion} and its derivatives, are typically based on volumetric integration of the Truncated Signed Distance Field (TSDF). These approaches incrementally fuse sequential depth maps to update a volumetric grid, from which a triangle mesh is subsequently extracted by Marching Cubes~\cite{lorensen1987marching} at a much lower frequency than the TSDF fusion. However, they are often constrained by fixed voxel resolution and high memory consumption, which prevent detailed capture and large-scale scene reconstruction.

\begin{figure}[t]
      \centering
      \includegraphics[width=\linewidth]{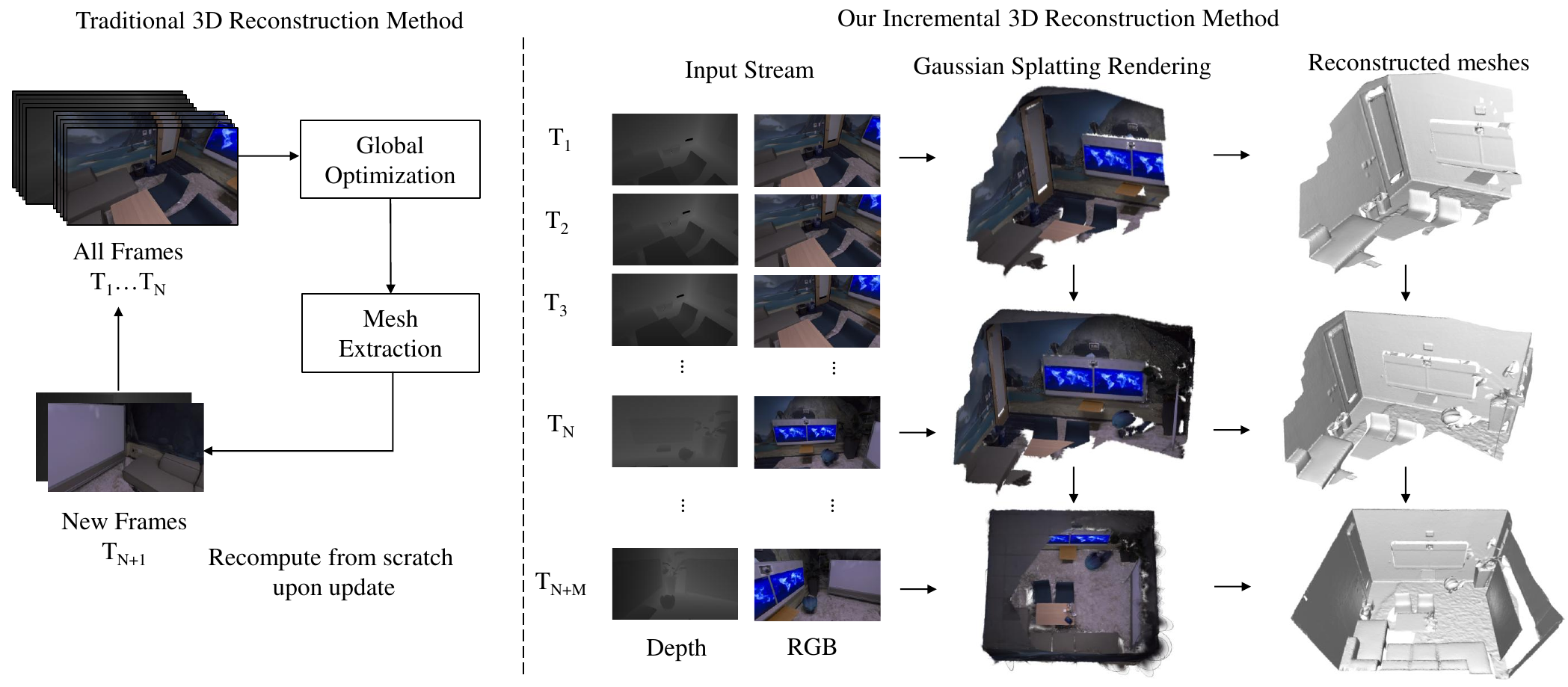}
      \caption{Comparison between traditional batch reconstruction and our proposed incremental framework. Left: Traditional methods typically rely on global optimization over all available frames ($T_1...T_N$). Once a new frame ($T_{N+1}$) is obtained, the entire mesh extraction pipeline has to be recomputed from scratch. Right: Our approach performs in a fully incremental manner. As the input stream arrives ($T_1 \dots T_{N+M}$), we progressively update the Gaussian representation and the reconstructed mesh. This allows for continuous scene expansion without global re-optimization. 
      }
      \label{fig:incremental}
\end{figure}

A remedy is to utilize continuous neural implicit representations, notably Neural Radiance Fields (NeRF)~\cite{mildenhall2021nerf}, which models scenes as radiance fields with adaptive sampling. Through spatial interpolation~\cite{sucar2021imap}  and feature priors~\cite{zhu2022nice}, these implicit methods can effectively reduce reconstruction artifacts and fill in missing geometry due to noisy input. However, these methods introduce new challenges, including heavy load during training and rendering due to computationally intensive network queries, and limited scalability in unbounded scenes. Furthermore, extracting explicit meshes from NeRF still requires dense sampling and offline post-processing, which hinders it from incremental scenarios.

To deal with the challenges of reconstruction and rendering speed, 3D Gaussian Splatting (3DGS) provides a notable advance in real-time novel view synthesis through its explicit and differentiable representation, though its potential in mesh reconstruction is overlooked. Existing methods like SuGaR~\cite{guedon2024sugar} and 2D Gaussian Splatting~\cite{huang20242d} attempt to address this by facilitating surface-aligned regularization and explicit 2D planar Gaussian representations to represent surfaces. 
In general, these methods decouple mesh extraction from the Gaussian representation and rely on global Poisson reconstruction~\cite{kazhdan2006poisson} or Marching Cubes~\cite{lorensen1987marching}, which require offline processing of the fully optimized Gaussian set.

In contrast to the conventional offline holistic methods, we propose a novel online incremental scene reconstruction framework that progressively integrates incoming data to generate triangular meshes. Our key idea is to treat optimized 3D Gaussians as surfel primitives for direct triangulation. By directly extracting watertight meshes from these primitives, our method leverages the flexibility of the explicit Gaussian representation to correct noisy depth regions, achieving fast and accurate reconstruction and rendering. 
To ensure these primitives are mesh-ready, we introduce: (i) a geometric planar constraint that aligns Gaussians to local surfaces, and (ii) a sparsity regularization that eliminates outliers while maintaining triangulation-conforming density. Furthermore, by freezing fully optimized mesh regions, our method prevents unbounded resource growth during long-sequence scanning. As illustrated in \cref{fig:incremental}, this enables scalable, continuous surface reconstruction.

The main contributions of this paper are as follows:
1) a novel triangulation algorithm on the Gaussian set, which achieves fast mesh extraction and incremental update from Gaussian surfels; 
2) a plane-based pulling constraint that guides Gaussian surfels to form a coherent structure in order to rectify input noise and faithfully represent the underlying surfaces;
3) a memory-efficient incremental scene reconstruction method that progressively freezes optimized mesh regions and focuses optimization on newly observed areas.

\section{Related Work}
\subsection{Novel View Synthesis and Gaussian Splatting}
We give a brief overview of Novel View Synthesis (NVS) methods for scene reconstruction. Neural Radiance Field (NeRF)~\cite{mildenhall2021nerf} and its variants~\cite{barron2021mip} have shown remarkable capabilities in synthesizing impressive novel views from multi-view images. However, the volumetric rendering is computationally intensive, which severely limits its efficiency. In contrast to NeRF, 3D Gaussian Splatting~\cite{kerbl20233d} is an explicit scene representation, which offers a more efficient alternative for novel view synthesis. Some subsequent works~\cite{yu2024mip, lu2024scaffold} have improved rendering quality. Most of them depend on the sparse point clouds generated by Structure-from-Motion (SfM) as initial input, which limits their applicability. To produce more
compact and accurate 3D Gaussians, GaussianPro~\cite{cheng2024gaussianpro} presents a novel progressive propagation strategy. GaMeS~\cite{waczynska2024games} takes the explicit mesh as input and parameterizes Gaussian ellipsoids with triangle meshes. Using sparse SfM point clouds or meshes as initialization for 3D Gaussian Splatting typically requires additional preprocessing steps and may introduce certain limitations. For large-scale scenes, RTG-SLAM~\cite{peng2024rtg} proposes a compact Gaussian representation and on-the-fly Gaussian optimization to facilitate real-time 3D reconstruction and superior renderings. Despite the fact that RTG-SLAM is able to incrementally reconstruct a Gaussian map, it still requires additional data structures and routines to obtain the reconstructed mesh. Our method not only synthesizes high-quality novel views, but also achieves incremental surface reconstruction.

\subsection{Surface Reconstruction from Gaussians}
Although 3D Gaussian Splatting has demonstrated remarkable NVS performance, its discrete and unstructured nature poses challenges for surface reconstruction.
SuGaR~\cite{guedon2024sugar} regularizes the Gaussians to be flat and well distributed
over the scene surface and derives a volume density from the Gaussians. To achieve a close surface alignment with Gaussians, 2D GS~\cite{huang20242d} and Gaussian surfels~\cite{dai2024high} employ surfels as surface representation. This involves conceptually flattening 3D Gaussian points into 2D by setting their z-scale to zero.
GS-Pull~\cite{zhang2024neural} aligns 3D Gaussians on the zero level-set of the neural SDF using neural pulling and updates the neural SDF through the pulled 3D Gaussians. These methods usually rely on either Poisson reconstruction, derived from an implicit field generated by Gaussians, or TSDF fusion using multi-view depth maps. Moreover, these surface extraction methods are constrained by high memory usage for large scenes. Gaussian Opacity Fields~\cite{yu2024gaussian} infers minimal opacity across views to keep multi-view consistency, followed by surface extraction using Marching Tetrahedra. The computation of the opacity values using all training views may lead to redundant computations.
Without the construction and maintenance of implicit fields, our proposed method directly obtains the detailed surface reconstruction from a Gaussian set and yields photorealistic rendering results.

\section{Method}
\subsection{Overview}
\begin{figure*}[t]
  \centering
  \includegraphics[width=\linewidth]{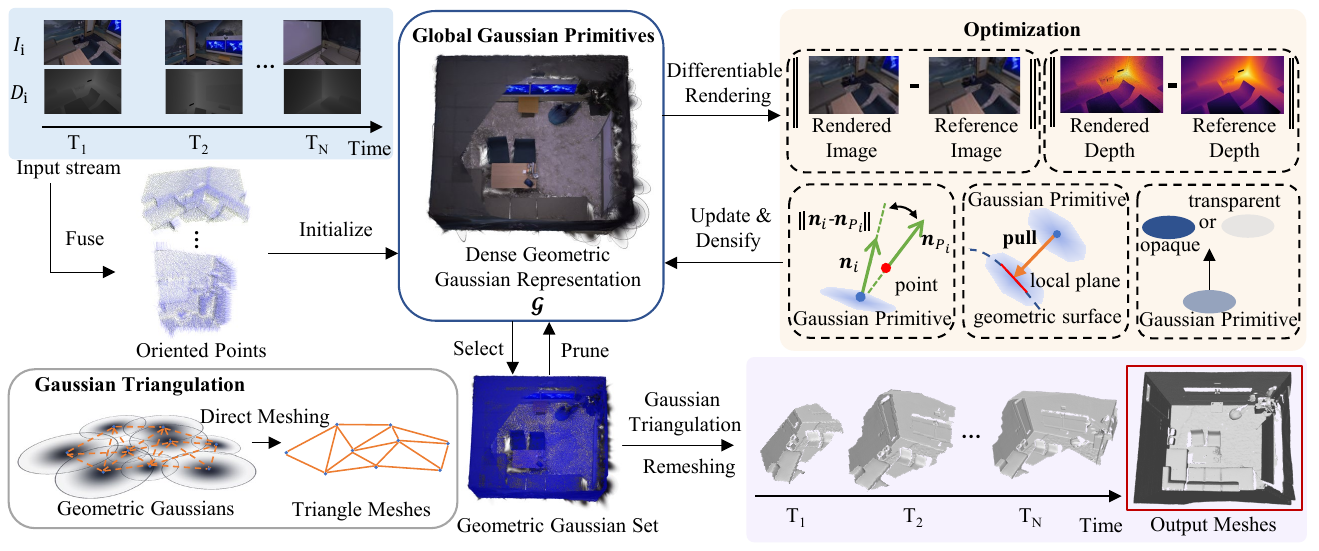}
  \caption{Overview of our proposed approach. We first initialize Gaussians from a continuous stream of oriented point clouds. The dense geometric Gaussian representation is progressively refined by loss-guided densification and optimization against depth, color, and geometric constraints. Next, a geometric Gaussian set is selected for meshing, during which process redundant primitives are pruned to maintain compactness. A novel triangulation algorithm (illustrated in the bottom-left inset) is subsequently applied to the selected Gaussians to incrementally reconstruct and update meshes.
  }
  \label{fig:overview}
\end{figure*}

We present a unified framework for incremental mesh reconstruction and high-fidelity rendering, as illustrated in \cref{fig:overview}. Our method takes a sequence of $M$ RGB-D frames $\{I_i, D_i\}_{i=1}^M$ with known camera poses as input. The foundation of our method is the Dense Geometric Gaussian representation, denoted as $\mathcal{G}$, which serves as the fundamental primitive for both volumetric rendering and explicit surface extraction (\cref{section:dffRepresentation}). To achieve incremental meshing from Gaussians, dense geometric Gaussian representation is dynamically maintained and progressively optimized under photometric and structural constraints. The combined supervision enforces the optimized Gaussians in order to adhere to the underlying surface geometry (\cref{section:onlinemapping}). At each timestamp, local meshes are obtained by triangulating a selected Geometric Gaussian Set. To ensure a memory-efficient incremental scene reconstruction, our method progressively freezes these well-optimized mesh regions and focuses optimization solely on newly observed areas (\cref{section:triangulation}). 

\subsection{Dense Geometric Gaussian Representation}\label{section:dffRepresentation}
\textbf{Geometric Gaussian Primitives.}
Following~\cite{kerbl20233d}, we represent the scene using a set of 3D Gaussian primitives $\{G_i\}$.
Each Gaussian comprises a 3D mean $\boldsymbol{\mu}_i\in \mathbb{R}^{3}$, opacity value $\alpha_i\in[0,1]$, RGB color values $\mathbf{c}_i\in \mathbb{R}^{3}$, scale vector $\mathbf{s}_i=[s_1^i,s_2^i,s_3^i]^\top\in\mathbb{R}^3$ and rotation $\mathbf{R}_i=[\mathbf{r}_1^i~\mathbf{r}_2^i~\mathbf{r}_3^i]\in \mathbb{R}^{3\times3}$.  

Unlike standard 3DGS optimized for volumetric radiance fields, our objective is high-fidelity surface reconstruction.
To this end, we explicitly constrain the Gaussian primitives to act as Planar Elliptical Surfels:
\begin{itemize}
\item \textbf{Planar Constraint:} We enforce the third scale component $s_3^i \to 0$, effectively flattening the Gaussian to align with the local tangent plane of the surface.
\item \textbf{Opacity Constraint:} To approximate hard physical geometry, we enforce high opacity ($\alpha_i \approx 1$).
\end{itemize}
These constraints allow our representation to seamlessly unify mesh reconstruction and rendering within a single framework.

\noindent\textbf{Gaussian Initialization.}
We initialize the geometric Gaussians from geometric attributes of points in the oriented point cloud $\mathcal{P}$, which is incrementally constructed from RGB-D streams. An oriented point in $\mathcal{P}$ has a position $\mathbf{p}_i$ and normal $\mathbf{n}_i$. The mean $\boldsymbol{\mu}_i$ is directly inherited from $\mathbf{p}_i$. Following~\cite{jiang2024gaussianshader}, we construct the rotation matrix $\mathbf{R}_i$ such that its third column $\mathbf{r}_3^i$ aligns with the normal vector $\mathbf{n}_i$.
To ensure proper surface coverage, the tangential scales ($s_1, s_2$) are determined by the local point spacing:
\begin{equation}
s_1^i = s_2^i = \lambda \cdot \max_{j \in \mathcal{N}(i)} \|\mathbf{p}_i - \mathbf{p}_j\|_2,
\label{equ:radius}
\end{equation}
where $\mathcal{N}(i)$ denotes the $k$-nearest neighbors in 3D space, $k=4$ and $\lambda=1.5$. Consistent with planar elliptical surfels, we initialize $s_3^i = 10^{-6}$ and $\alpha_i = 0.99$.

\noindent\textbf{Geometry-Aware Differentiable Rendering.}
We adopt a dual-branch rendering strategy to optimize both photometric and geometric consistency. We render the optimizable Gaussians $\{G_i\}$ into RGB images $\hat{\mathbf{C}}$ through the standard alpha-blending proposed in \cite{kerbl20233d},
\begin{equation}
    \hat{\mathbf{C}}(u,v)=\sum_{i\in N} \mathbf{c}_i\alpha_i \prod_{j=1}^{i-1}(1-\alpha_j),
    \label{equ:alphablending}
\end{equation}
where $\mathbf{c}_i$ denotes the Gaussian color, and $\alpha_i$ is its opacity.

The standard alpha-blending depth is insufficient for precise mesh reconstruction. To model the local geometric plane as opaque Gaussian primitives, we compute the intersection of the view ray and the front opaque Gaussian primitive to obtain the depth at a pixel $\hat{\mathbf{D}}(u,v)$, as in~\cite{peng2024rtg}. 
The depth map $\hat{\mathbf{D}}$ is computed as below 
\begin{equation}\label{equ:depthmap}
\hat{\mathbf{D}}(u,v)=
\begin{cases}
         -1 \quad  \text{if no intersection with opaque Gaussians}, \\
         (\mathbf{T}^c_w (\frac{\boldsymbol{\mu}_j^\mathbf{r}\cdot \mathbf{n}_j^\mathbf{r}}{\mathbf{r}\cdot \mathbf{n}_j^\mathbf{r}} \cdot \mathbf{r} ))_z \quad  \text{if}\; \langle\mathbf{n}_j^\mathbf{r}, \mathbf{r}\rangle < 0.5,\\
         (\mathbf{T}^c_w \boldsymbol{\mu}_j^\mathbf{r})_z \quad \text{otherwise}.
\end{cases}
\end{equation}
where $\mathbf{r}$ denotes the normalized ray in the world coordinate. $\boldsymbol{\mu}_j^\mathbf{r}$ and $\mathbf{n}_j^\mathbf{r}$ are the position and normal of the intersected Gaussian, respectively. $\mathbf{T}^c_w$ represents the transformation from the world coordinate to the current camera, and $(\cdot)_z$ extracts the $z$-component of the 3D vector.

\subsection{Online Mapping} \label{section:onlinemapping}
\textbf{Gaussian Adding.}
Based on the newly added oriented point cloud, we first add corresponding Gaussians. 
To compensate for the information loss prone to insufficient depth observations, we dynamically supplement missing regions driven by both geometric and photometric rendering discrepancies. For significant depth differences ($\mathbf{D}_i - \hat{\mathbf{D}}_i$), we add opaque Gaussians to improve the completeness of the reconstructed mesh. Similarly, we leverage the RGB difference $\Delta \mathbf{C}_i = \mathbf{C}_i - \hat{\mathbf{C}}_i$ to add transparent Gaussians (set $\alpha_i = 0.1$ ) so that the rendering quality can be greatly enhanced. 

\noindent\textbf{Gaussian Pruning.}
Since our representation relies on opaque Gaussians to fit local surfaces, we prune opaque Gaussians that do not participate in depth rendering and transparent Gaussians that contribute minimally to photometric rendering. The pruning operates concurrently with Gaussian optimization process, incurring negligible computational overhead.

\noindent\textbf{Gaussian Optimization.}\label{section:gaussian_optimization}
After adding Gaussians for the active window, we iteratively optimize their attributes by randomly sampling one frame per iteration. To ensure accurate reconstruction and rendering, we supervise this process with a combination of photometric and structural constraints. 

We formulate the rendering losses to supervise the Gaussian optimization:
\begin{equation}
    \mathcal{L}_{color} = \sum_{i}^N|\mathbf{C}_i - \hat{\mathbf{C}}_i|, \quad \mathcal{L}_{depth} = \sum_{i}^N|\mathbf{D}_i - \hat{\mathbf{D}}_i|.
\end{equation}
By leveraging the complementary nature of photometric and geometric constraints, Gaussian surfels can robustly recover scene geometry, mutually compensating for sensor noise and depth missing.

To enable the direct reconstruction of smooth and accurate surfaces from Gaussian sets, we introduce two geometric constraints. The first constraint aims to align the extracted Gaussians with the underlying local geometry. Inspired by GS-Pull~\cite{zhang2024neural}, we consider our oriented point clouds as a discrete approximation of the surface zero level set $Z(f)$. Accordingly, we introduce a local plane loss to pull Gaussian centers $\boldsymbol{\mu}_i$ toward the underlying surface points represented by the oriented point clouds as follows
\begin{equation}
\mathcal{L}_{plane} = \sum_{(\boldsymbol{\mu}_i, \mathbf{p}_i) \in \mathcal{C}} |\mathbf{n}^\top_i(\boldsymbol{\mu}_i - \mathbf{p}_i)|  ,
    \label{equ:planarloss}
\end{equation}
where $\mathcal{C}$ is the set of correspondences between geometric Gaussians and oriented points, $\mathbf{p}_i \in \mathcal{P}$ is the point position, and $\mathbf{n}_i$ is the point normal. $\mathcal{P}$ denotes oriented point clouds. The plane-based pulling loss encourages the Gaussian centers to lie on the local plane while preserving maximal freedom during optimization.

Our second geometric constraint enforces the orientational alignment between the Gaussians and the local surfaces.
Accordingly, we employ a normal consistency loss defined as
\begin{equation}
\mathcal{L}_n = 1 - |\mathbf{n}^\top_{g} \cdot \mathbf{n}_i|  ,
    \label{equ:normal}
\end{equation}
where $\mathbf{n}_g$ is the Gaussian normal, which is approximated by the shortest-axis direction of each Gaussian surfel following GaussianShader~\cite{jiang2024gaussianshader}.

To facilitate photometric rendering and geometric optimization, we utilize a sparse loss~\cite{jiang2024gaussianshader} to encourage Gaussians to become either fully transparent or fully opaque.
\begin{equation}\label{equ:sparseloss}
    \mathcal{L}_{\text{sparse}} = \frac{1}{|\alpha|} \sum_{\alpha_i} \left[ \log(\alpha_i) + \log(1 - \alpha_i) \right].
\end{equation}

Above all, we optimize the attributes of Gaussians via the following objective function, 
\begin{equation} \label{total_loss}
    \mathcal{L}=\lambda_c\mathcal{L}_{color} + \lambda_d\mathcal{L}_{depth} + \lambda_{p}\mathcal{L}_{plane} +  \lambda_{n}\mathcal{L}_{n} + \lambda_{s}\mathcal{L}_{sparse}.
\end{equation}

\subsection{Gaussian Triangulation} \label{section:triangulation}
\textbf{Geometric Gaussian Set.} 
To enable accurate meshing from dense geometric Gaussian representation, we need to filter out Gaussian primitives that serve merely for visual rendering rather than for true geometry. Therefore, we propose a simple and effective strategy that relies on the opacity and depth fidelity to extract the Geometric Gaussian Set $G_{geo}$, a subset of $\{G_i\}$. Specifically, the geometric Gaussian set $G_{geo}$ is defined as
\begin{equation}
     G_{geo} = \{G_g \in \{G_i\} \mid \alpha_g > 0.9 \land  (|(\mathbf{T}_{cw} \mathbf{\mu}_g)_z - D| < \tau) \}.
     \label{equ:gaussiansurfel}
\end{equation}
where $\tau$ denotes a predefined threshold. The opacity criterion removes semi-transparent Gaussians that model volumetric or view-dependent artifacts, while the depth fidelity criterion eliminates geometrically inaccurate primitives. This strategy ensures that only robust, highly opaque primitives corresponding to actual physical structures are retained to reconstruct the explicit meshes.

\begin{figure*}[t]
\centering
    \centerline{\includegraphics[width=0.98\textwidth]{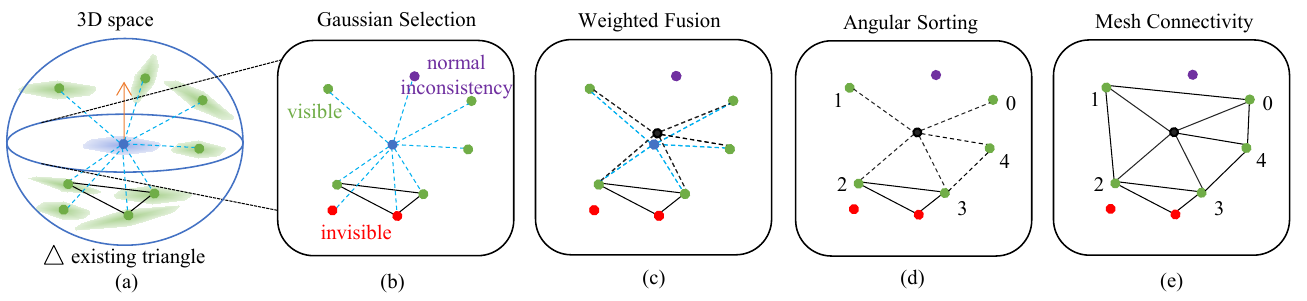}} 
    \caption{Illustration of Gaussian triangulation pipeline. 
    (a) Neighbors within an adaptive radius are identified and projected onto the tangent plane. (b) Valid candidates (green) are retained, while invisible (red) and normal-inconsistent (purple) ones are pruned. (c) The central Gaussian (blue) is shifted to a refined position (black) via weighted fusion. (d) Neighbors are re-projected onto the new tangent space and sorted angularly (indexed 0-4). (e) Mesh connectivity is established via an angle-based greedy algorithm. 
    }
    \label{fig:overview_meshing}

\end{figure*}

\noindent\textbf{Fast Gaussian Triangulation.} 
Inspired by fast point set triangulation~\cite{gopi2000fast,marton2009fast}, we propose a triangulation approach for the geometric Gaussian set, as illustrated in \cref{fig:overview_meshing}. To ensure computational efficiency, we employ a compressed octree.
For a given Gaussian $G_{i}$, an adaptive search radius is defined based on the mean distance to its $k$-nearest neighbors ($k=4$ empirically). Within this radius, we identify a valid neighbor subset $\mathcal{N}_{G}$ by enforcing two critical geometric constraints on each candidate $G_j$: mutual visibility on the tangent plane and normal consistency ($\langle \mathbf{n}_i, \mathbf{n}_j \rangle > 0.9$).
To yield a more coherent surface, the 3D mean of each Gaussian is refined through a weighted averaging of these candidates. Following the global update of the Gaussian set, we perform a second neighbor search on the refined Gaussians to establish the final connectivity for triangulation.

To establish the topological connectivity, we project all candidate neighbors onto the tangent plane of the central Gaussian. In this 2D local frame, the central Gaussian acts as the pivot vertex. We employ an angle-based greedy strategy to reconstruct the triangular meshes: neighbors are first sorted by polar angle around the center. We then iteratively form triangles by connecting the central Gaussian with pairs of angularly adjacent neighbors. To ensure mesh quality and prevent the formation of degenerate sliver triangles, we enforce a constraint that the subtended angle must exceed $10^{\circ}$.

\noindent\textbf{Local Remeshing and Freezing.} 
In incremental scene reconstruction, the system generates a local triangular mesh for the current viewpoint at timestamp $t$ and attempts to fuse it with the existing surface mesh. Due to observational variations across different viewpoints, direct integration inevitably leads to vertex misalignments and topological conflicts in adjacent regions. To address this issue, we introduce a local remeshing strategy to eliminate topological conflicts and achieve seamless stitching. Following the isotropic mesh optimization approach proposed by Botsch et al.~\cite{botsch2004remeshing}, we begin by splitting edges longer than $1.5\bar{L}$ (where $\bar{L}$ is the local average edge length). Subsequently, edges shorter than $0.5\bar{L}$ are collapsed to simplify the topology. Then, edges are evaluated and flipped if it optimizes the degree of neighboring vertices towards the target value of 6. Finally, vertices are relocated to the centroid of their neighbors via Laplacian smoothing. We repeat this process for three iterations to enforce strict geometric consistency between the newly observed region and the prior mesh.

Furthermore, to prevent the unbounded growth of computational and memory overhead associated with long-sequence reconstruction, we freeze the triangular meshes and Gaussian surfels in regions that possess extensive historical observations and are fully optimized. Specifically, the system quantifies the convergence state of a region by evaluating the multi-view observation counts and the loss of its Gaussians. When the cumulative effective observation count of Gaussians within a local region exceeds a predefined threshold $N_{obs}$, and their corresponding optimization loss $\mathcal{L}$ has stably converged to a minimum $\epsilon_{gs}$, the region is determined to be fully optimized. Once this condition is met, the system removes the Gaussian surfels and mesh of this region from the active optimization computation graph and freezes them. Such combined strategy of dynamically freezing stable regions alongside local remeshing ensures that the system maintains excellent memory efficiency when processing long-sequence scenes.

\section{Experiments}
In this section, we present the details of our experiments and conduct experiments on public datasets to evaluate the reconstruction accuracy and rendering quality of our proposed approach.

\subsection{Experimental Setup}
\noindent\textbf{Implementation Details.} Our framework is implemented in Python, with the Gaussian set triangulation and oriented point cloud generation encapsulated in C++ modules. We leverage CUDA parallel computing to accelerate the generation of oriented point clouds. For Gaussian optimization, we set $\lambda_c=0.8, \lambda_d=1.0, \lambda_{p}=0.05, \lambda_{n}=0.2, \lambda_{s}=0.001$. For Gaussian filtering, we use $\tau =0.001$ to select the geometric Gaussian set. Additionally, the observation count threshold $N_{obs}$ is set to 10, and the geometric selection loss threshold $\epsilon_{gs}$ is set to 0.1. The active window size and optimization iterations are set to $6$ and $50$ for Replica~\cite{straub2019replica}, and $3$ and $75$ for ScanNet++~\cite{yeshwanthliu2023scannetpp}, respectively. All experiments were conducted on a PC equipped with an NVIDIA RTX 3090 GPU. To facilitate a fair evaluation of reconstruction quality and rendering performance, we utilize ground truth poses for all baselines in our experiments.

\noindent\textbf{Datasets.} We mainly evaluate the proposed method on two datasets, namely Replica~\cite{straub2019replica} and ScanNet++~\cite{yeshwanthliu2023scannetpp}. For Replica, we use eight sequences featuring living room and office scenarios.  For ScanNet++, we use the DSLR captures from two scenes (8b5caf3398 (S1) and b20a261fdf (S2)).

  \begin{figure*}[t]
    \centering
        \centerline{\includegraphics[width=\textwidth]{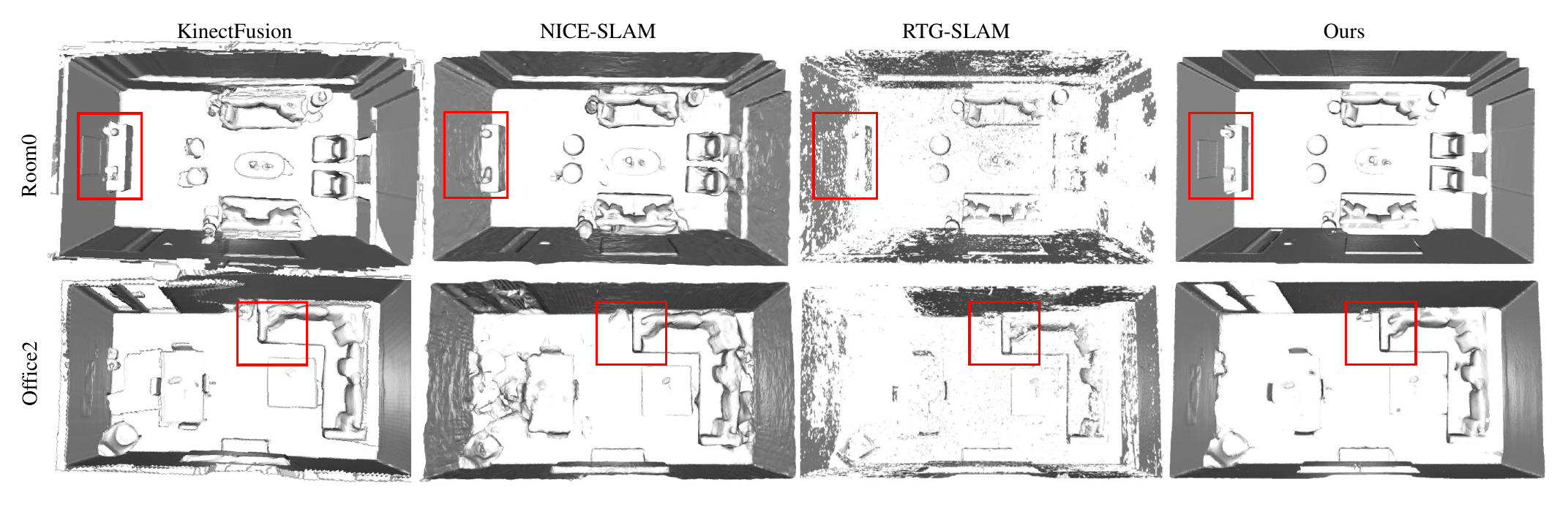}} 
    	\caption{Comparison of Mesh Reconstruction results across different Replica sequences. For RTG-SLAM~\cite{peng2024rtg}, we employ the ball pivoting algorithm~\cite{bernardini2002ball} to generate meshes from Gaussian balls.
     }
    	\label{fig:meshing_com}
    
   \end{figure*}

\begin{table}[t]
    \centering
    \caption{Comparison of geometry accuracy on Replica.}
    \tabcolsep=0.15cm
    \setlength{\tabcolsep}{1.3pt}

    \resizebox{\linewidth}{!}{
\begin{tabular}{c c c c c c c c c c c}
\toprule
   Method     & Metric               & Rm 0   & Rm 1  & Rm 2  & Off 0 & Off 1 & Off 2 & Off 3 & Off 4 & Avg. \\ \hline
\multirow{3}{*}{KinectFusion~\cite{newcombe2011kinectfusion}} & Acc.[cm]$\downarrow$  & 10.39 & 19.27 & 17.25 & 11.10 & 9.99 & 13.43 & 9.88 & 10.77 & 12.76 \\
        & Acc. Ratio$\uparrow$  & 44.02 & 22.36 & 24.68 & 37.13 & 36.64 & 34.31 & 39.58 & 35.98 & 34.34 \\
        & Comp. Ratio$\uparrow$  & 79.20 & 47.23 & 50.98 & 72.25 & 70.57 & 60.93 & 62.26 & 59.30 & 62.84 \\ \hline

\multirow{3}{*}{NICE-SLAM~\cite{zhu2022nice}} & Acc.[cm]$\downarrow$  & 3.22 & 42.36 & 2.61 & 2.32 & 3.22 & 3.85 & 4.70 & 2.99 & 8.16 \\
        & Acc. Ratio$\uparrow$  & 89.33 & 39.01 & 91.77 & 93.39 & 87.02 & 83.26 & 68.12 & 91.44 & 80.42 \\
        & Comp. Ratio$\uparrow$  & \textbf{88.39} & 73.80 & \textbf{90.73} & \textbf{94.80} & \textbf{94.38} & \textbf{84.12} & 67.16 & \textbf{86.71} & 85.01 \\ \hline

\multirow{3}{*}{MonoGS~\cite{matsuki2024gaussian}} & Acc.[cm]$\downarrow$  & 3.26 & 3.23 & 2.89 & 2.32 & 1.93 & 2.76 & 2.72 & 3.02 & 2.77 \\
        & Acc. Ratio$\uparrow$  & 82.83 & 81.95 & 85.97 & 90.01 & 91.45 & 87.10 & 89.05 & 85.46 & 86.73 \\
        & Comp. Ratio$\uparrow$  & 80.45 & 82.59 & 82.45 & 91.39 & 86.91 & 76.04 & 75.44 & 79.33 & 81.83 \\ \hline
\multirow{3}{*}{RTG-SLAM~\cite{peng2024rtg}}          & Acc.[cm]$\downarrow$      & 1.67 & 1.23 & 1.32 & 1.16 & 0.92 & 1.48 & 1.80 & 1.67 & 1.41 \\
        & Acc. Ratio$\uparrow$  & 99.16 & \textbf{99.90} & \textbf{99.74} & 99.91 & 99.86 & 99.40 & 98.85 & 99.54 & 99.54 \\
        & Comp. Ratio$\uparrow$ & 78.80 & 83.97 & 84.87 & 88.40 & 87.66 & 76.79 & 74.51 & 78.68 & 81.71 \\ \hline
\multirow{3}{*}{Ours}         & Acc.[cm]$\downarrow$  & \textbf{1.52} & \textbf{1.20} & \textbf{1.29} & \textbf{1.09} & \textbf{0.88} & \textbf{1.44} & \textbf{1.78} & \textbf{1.54} & \textbf{1.34} \\
 & Acc. Ratio$\uparrow$  & \textbf{99.85} & 99.85 & 99.53 & \textbf{99.93} & \textbf{99.99} & \textbf{99.67} & \textbf{99.05} & \textbf{99.73} & \textbf{99.70} \\
 & Comp. Ratio$\uparrow$ & 85.24 & \textbf{86.35} & 85.08 & 91.53 & 88.32 & 83.71 & \textbf{82.84} & 84.51 & \textbf{85.95} \\
        
        \bottomrule
\end{tabular}
}
 \label{tab:geometry_accuracy_replica}
\end{table}

\begin{table}[t]
    \centering
    \caption{Comparison of rendering performance on Replica~\cite{straub2019replica}.}
    \tabcolsep=0.15cm
    \setlength{\tabcolsep}{1.2pt}

    \resizebox{\linewidth}{!}{
    \begin{tabular}{c l c c c c c c c c c}
    \toprule
        Method     & Metric               & Rm 0   & Rm 1  & Rm 2  & Off 0 & Off 1 & Off 2 & Off 3 & Off 4 & Avg. \\ \hline
 \multirow{3}{*}{NICE-SLAM~\cite{zhu2022nice}} 
                        &PSNR$\uparrow$      & 24.72  & 26.79 & 27.06 & 30.21 & 32.78 & 26.59 & 26.22 & 24.74 & 27.39 \\
                        &SSIM$\uparrow$      & 0.787   & 0.799  & 0.807  & 0.881  & 0.906  & 0.816  & 0.801  & 0.834  & 0.829 \\
                        &LPIPS$\downarrow$   & 0.431   & 0.372  & 0.329  & 0.322  & 0.275  & 0.321  & 0.288  & 0.333 & 0.334 \\ \hline
    \multirow{3}{*}{Point-SLAM~\cite{sandstrom2023point}} & PSNR$\uparrow$    & 32.40          & 34.08          & 35.50          & 38.26          & 39.16          & 33.99          & 33.48          & 33.49          & 35.17          \\
        & SSIM$\uparrow$    & 0.97           & 0.98           & 0.98           & 0.98           & \textbf{0.99}  & 0.96           & 0.96           & 0.98           & 0.98           \\
        & LPIPS$\downarrow$ & 0.11           & 0.12           & 0.11           & 0.10           & 0.12           & 0.16           & 0.13           & 0.14           & 0.12           \\ \hline
   
    \multirow{3}{*}{MonoGS~\cite{matsuki2024gaussian}}    & PSNR$\uparrow$    & 34.83          & 36.43          & 37.49          & 39.95          & 42.09          & \textbf{36.24} & \textbf{36.70} & 36.07          & 37.50          \\
        & SSIM$\uparrow$    & 0.95           & 0.96           & 0.97           & 0.97           & 0.98           & 0.96           & 0.96           & 0.96           & 0.96           \\
        & LPIPS$\downarrow$ & 0.07           & 0.08           & 0.08           & 0.07           & 0.06           & 0.08           & 0.07           & 0.10           & 0.07           \\ \hline
    \multirow{3}{*}{RTG-SLAM~\cite{peng2024rtg}}          & PSNR$\uparrow$    & 31.56          & 34.21          & 35.57          & 39.11          & 40.27          & 33.54          & 32.76          & 36.48          & 35.43          \\
        & SSIM$\uparrow$    & 0.97           & 0.98           & 0.98           & \textbf{0.99}  & \textbf{0.99}  & 0.98           & 0.98           & 0.98           & 0.98           \\
        & LPIPS$\downarrow$ & 0.13           & 0.11           & 0.12           & 0.07           & 0.08           & 0.13           & 0.13           & 0.12           & 0.11           \\ \hline
    \multirow{3}{*}{Ours}                                                       & PSNR$\uparrow$    & \textbf{35.12} & \textbf{36.58} & \textbf{37.61} & \textbf{42.65} & \textbf{42.43} & 35.30          & 34.86          & \textbf{38.27} & \textbf{37.85} \\
        & SSIM$\uparrow$    & \textbf{0.99}  & \textbf{0.99}  & \textbf{0.99}  & \textbf{0.99}  & \textbf{0.99}  & \textbf{0.99}  & \textbf{0.99}  & \textbf{0.99}  & \textbf{0.99}  \\
        & LPIPS$\downarrow$ & \textbf{0.04}  & \textbf{0.04}  & \textbf{0.03}  & \textbf{0.02}  & \textbf{0.02}  & \textbf{0.03}  & \textbf{0.05}  & \textbf{0.03}  & \textbf{0.03}  \\ 
    \bottomrule
    \end{tabular}
    }
    \label{tab:rendering_replica}
\end{table}

\begin{figure}[!t] 
\resizebox{\linewidth}{!}{
    \centering
    \begin{subfigure}[t]{0.535\linewidth}
        \centering
        \includegraphics[width=\linewidth]{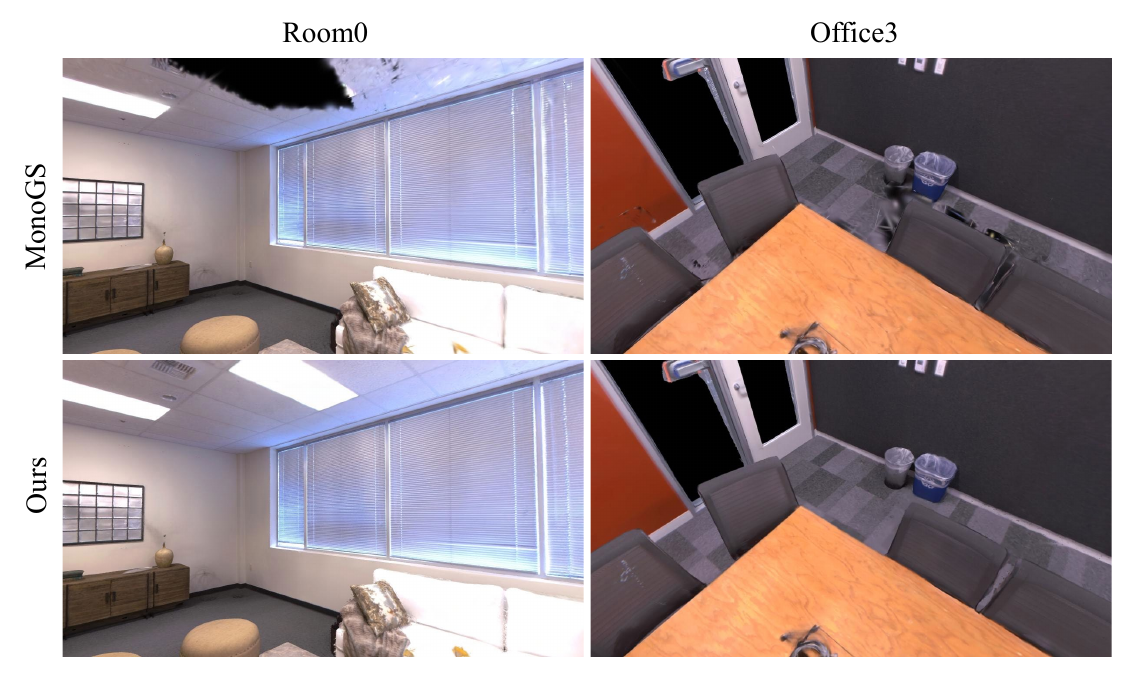}
        \caption{Comparison on Replica (vs. MonoGS)}
        \label{fig:rendering_replica}
    \end{subfigure}
    \hfill 
    \begin{subfigure}[t]{0.46\linewidth} 
        \centering
        \includegraphics[width=\linewidth]{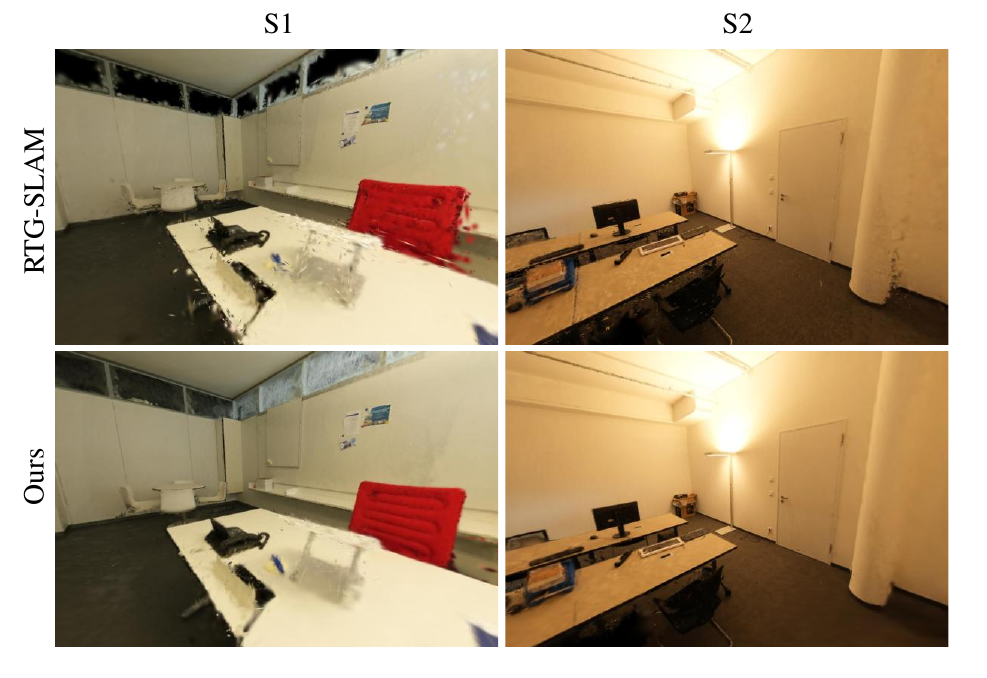}
        \caption{Comparison on ScanNet++ (vs. RTG-SLAM)}
        \label{fig:nvs_rendering_scannet++}
    \end{subfigure}
        }
    \caption{
        \textbf{Qualitative comparisons of novel view synthesis across different datasets.} 
        \textbf{(a)} Results on Replica compared with MonoGS. Our method reconstructs sharper geometry details with fewer artifacts.
        \textbf{(b)} Results on ScanNet++ compared with RTG-SLAM. 
        Note that our approach produces cleaner textures in complex indoor environments.
    }

\end{figure}

\subsection{Reconstruction Evaluation}
We evaluate the reconstruction quality of our method on the Replica dataset, comparing it against the volumetric TSDF method KinectFusion~\cite{newcombe2011kinectfusion}, the NeRF-based RGB-D SLAM method NICE-SLAM~\cite{zhu2022nice}, and recent Gaussian SLAM approaches, including RTG-SLAM ~\cite{peng2024rtg} and MonoGS~\cite{matsuki2024gaussian}.
Meshes for RTG-SLAM are generated using the ball pivoting algorithm~\cite{bernardini2002ball}, which effectively exploits the geometric attributes of its Gaussians. As illustrated in \cref{fig:meshing_com}, our method produces finer geometric details and better preserves the smoothness of planar surfaces, such as walls. To evaluate the quality of the reconstructed geometry, we adopt the standard metrics: Accuracy, Accuracy Ratio ($<5$cm), and Completion Ratio ($<5$cm). For MonoGS, we uniformly sample an equal number of points from the reconstructed Gaussians, while other baselines are directly evaluated on the generated meshes. As shown in \cref{tab:geometry_accuracy_replica}, our proposed method not only achieves the best average performance but also attains the highest accuracy on each Replica sequence. 

\begin{table}[t]
 \caption{Novel \& Train View Rendering Performance on ScanNet++~\cite{yeshwanthliu2023scannetpp}.} 

\centering
\setlength{\tabcolsep}{3.5pt}

\begin{tabular}{ccccccccc}
\toprule

\multirow{2}{*}{\textbf{Methods}} & \multirow{2}{*}{\textbf{Metrics}} & \multicolumn{3}{c}{\textbf{Novel View}} & \multicolumn{3}{c}{\textbf{Training View}} \\

 &  & \textbf{Avg.} & \texttt{S1} & \texttt{S2} & \textbf{Avg.} & \texttt{S1} & \texttt{S2} \\

\midrule

\multirow{3}{*}{Point-SLAM~\cite{sandstrom2023point}} & PSNR $\uparrow$ & 22.53 & 22.03 & 23.02 & 25.94 & 25.34 & 26.54 \\
& SSIM $\uparrow$  & 0.77 & 0.75 & 0.78 & 0.86 & 0.83 & 0.89 \\
& LPIPS $\downarrow$ & 0.47 & 0.47 & 0.46  & 0.32 & 0.36 & 0.27 \\

\hline

\multirow{3}{*}{RTG-SLAM~\cite{peng2024rtg}}  & PSNR $\uparrow$    & 21.65 & 19.01 & 24.29 & 22.29 & 19.73 & 24.85 \\
& SSIM $\uparrow$     & 0.83 & 0.81 & 0.84 & 0.88 & 0.87 & \textbf{0.90} \\
& LPIPS $\downarrow$    & 0.33 & 0.34 & \textbf{0.33} & 0.24 & 0.26 & \textbf{0.23} \\
\hline

\multirow{3}{*}{Ours} & PSNR $\uparrow$ & \textbf{24.40} & \textbf{23.78} & \textbf{25.03} & \textbf{26.89} & \textbf{26.65} & \textbf{27.14} \\
& SSIM $\uparrow$  & \textbf{0.85} & \textbf{0.85} & \textbf{0.85} & \textbf{0.91} & \textbf{0.91} & \textbf{0.90} \\
& LPIPS $\downarrow$ & \textbf{0.32} & \textbf{0.30} & 0.34 & \textbf{0.23} & \textbf{0.21} & 0.24 \\
\bottomrule
\end{tabular}
\label{tab:scannetpp}
\end{table}

\begin{table}[t]
\small
\setlength{\tabcolsep}{3.5pt}
\resizebox{0.96\linewidth}{!}{

\begin{minipage}{0.48\linewidth}
\caption{Comparison of runtime and peak memory usage on Replica (Off0).}
\label{tab:rendering_speed}
\begin{tabular}{lcc}
\toprule
Method & FPS $\uparrow$ & Mem. (MB) $\downarrow$ \\
\midrule
MonoGS & 1.48 & 5434 \\
RTG-SLAM & 3.65 & 2751 \\
Ours & \textbf{10.34} & \textbf{2325} \\
\bottomrule
\end{tabular}
\end{minipage}
\hspace{0.005\linewidth}
\begin{minipage}{0.48\linewidth}
\caption{Efficiency Comparison of Gaussian-to-Mesh Conversion.}
\label{tab:ablation_meshing_time}
\begin{tabular}{@{}lcc@{}}
\toprule
Method & Mesh Size (MB)  $\downarrow$ & Time (s) $\downarrow$ \\
\midrule
GOF (MC) & 964.95 & 444.26 \\
GOF (PSR) & 14.05 & 415.27 \\
Ours & \textbf{3.32} & \textbf{5.34} \\
\bottomrule
\end{tabular}
\end{minipage}

}
\end{table}

\subsection{Rendering Evaluation}
We quantitatively evaluate the rendering performance of our method on the Replica dataset and adopt PSNR, SSIM, and LPIPS as the evaluation metrics. To demonstrate the efficacy of our proposed approach, we compare our method with NeRF RGB-D SLAM methods like NICE-SLAM~\cite{zhu2022nice}, Point-SLAM~\cite{sandstrom2023point}, and two Gaussian SLAM methods, including MonoGS~\cite{matsuki2024gaussian} and RTG-SLAM~\cite{peng2024rtg}. The results are directly from the publications~\cite{peng2024rtg, matsuki2024gaussian}. As shown in \cref{tab:rendering_replica}, our approach achieves superior rendering performance over all baseline methods in almost all sequences. The novel view synthesis results in \cref{fig:rendering_replica} demonstrate that our approach effectively reduces artifacts and preserves fine-grained details while generating more complete scenes.

We further assess the rendering capability of our method through quantitative and qualitative evaluations on the ScanNet++ dataset.
The experiments for RTG-SLAM and Point-SLAM~\cite{sandstrom2023point} were conducted based on their official setup. \cref{tab:scannetpp} shows that our method achieves superior rendering quality over Point-SLAM and RTG-SLAM in both training views and novel views. Furthermore, the novel view rendering results in \cref{fig:nvs_rendering_scannet++} qualitatively demonstrate that our method achieves finer details with fewer artifacts. Both quantitative and qualitative results on Replica and ScanNet++ demonstrate that our approach surpasses baseline methods in detail preservation, scene completeness, and artifact reduction.

\subsection{Efficiency} 
To demonstrate the efficiency of our approach, we evaluate the whole mapping frame rate (FPS) and peak memory usage during mapping on the `Office0' sequence of Replica. We compare with MonoGS and RTG-SLAM, conducting experiments on the same workstation for fairness. As shown in \cref{tab:rendering_speed}, our method achieves the highest mapping speed at 10.34 FPS and the lowest peak memory of 2325 MB. This efficiency stems from our incremental mesh freezing strategy. By freezing the triangular meshes and Gaussian surfels within fully optimized historical regions, our system restricts active optimization to a limited local window, substantially reducing memory overhead and accelerating computation.

To investigate mesh extraction efficiency, we compare with Gaussian Opacity Field (GOF)~\cite{yu2024gaussian}, which reconstructs meshes via a volumetric opacity field followed by Marching Cubes (MC) or Poisson Surface Reconstruction (PSR). Since GOF requires excessive memory for a global implicit field, we restrict the comparison to a small subset of the `Office0' scene from Replica. \Cref{tab:ablation_meshing_time} shows our method takes only 5.34 seconds, much faster than GOF-MC (444.26 s) and GOF-PSR (415.27 s). This speedup comes from directly triangulating Gaussian surfels locally without constructing dense 3D voxel grids or evaluating global implicit fields. Regarding mesh size, GOF demands excessively massive meshes (e.g., 964.95 MB for GOF-MC) to capture fine details. In contrast, our planar constraints align surfels tightly to the true surface, allowing higher-fidelity geometry with a compact mesh.

\begin{table}[t]
\centering
\caption{Ablation studies on key modules and loss terms for mesh reconstruction.}
\label{tab:ablation_key_losses}
\footnotesize
\resizebox{0.98\linewidth}{!}{
\begin{tabular}{lccc@{\hskip 6pt}|@{\hskip 6pt}lccc}
\toprule
Key Modules & Acc.[cm]$\downarrow$ & Acc. Ratio$\uparrow$ & Comp. Ratio$\uparrow$ &
Loss Terms & Acc.[cm]$\downarrow$ & Acc. Ratio$\uparrow$ & Comp. Ratio$\uparrow$ \\
\midrule
w/o Planar Constraint      & 1.19 & 99.04 & 88.58 &
Baseline                   & 1.13 & 98.58 & 86.34 \\
w/o Neighborhood Selection & 1.26 & 98.59 & 85.14 &
$+L_{sparse}$              & 1.09 & 98.92 & 86.50 \\
w/o Direct Triangulation   & 0.89 & 99.98 & 84.49 &
$+L_n+L_{sparse}$          & 0.95 & 98.72 & 86.13 \\
w/o Local Remeshing        & 1.32 & 97.99 & \textbf{91.10} &
$+L_{plane}+L_{sparse}$    & 0.93 & 99.78 & 86.50 \\
w/o Freezing               & 0.92 & 99.59 & 89.27 &
$+L_{plane}+L_n$           & 0.90 & 99.77 & 86.04 \\
Full Method         & \textbf{0.88} & \textbf{99.99} & 88.32 &
Full losses                & \textbf{0.88} & \textbf{99.99} & \textbf{88.32} \\
\bottomrule
\end{tabular}
}
\end{table}

  \begin{figure}[t]
    \centering
        \centerline{\includegraphics[width=\textwidth]{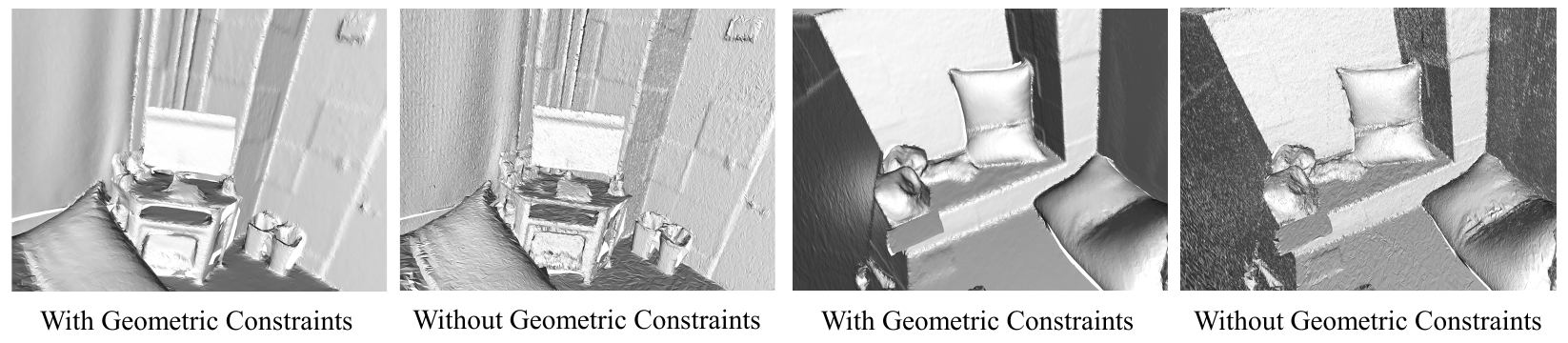}} 
    	\caption{Qualitative effect of geometric constraints on mesh reconstruction. We display the reconstruction details. When only depth constraints are applied without geometric constraints, the reconstruction captures the general outline of the scene but suffers from low surface fidelity and exhibits noticeably coarse textures.
     }
    	\label{fig:meshing_geometric_constraints}
    
   \end{figure}

\subsection{Ablation Study}
\textbf{Core Components.} \cref{tab:ablation_key_losses} reports ablation results on Replica (Office1). The left part evaluates key modules, while the right part studies the contribution of optimization loss terms. In `w/o Direct Triangulation', we adopt Poisson surface reconstruction in place of the proposed direct triangulation. Each module improves reconstruction quality, and the full model achieves the best Acc.\ and Acc.\ ratio.  All loss terms contribute positively, and the complete objective yields the best overall performance, reducing the accuracy error from 1.13 cm to 0.88 cm and improving the completion ratio from 86.34\% to 88.32\%.

\noindent\textbf{Geometric Constraints.}
To further validate the two geometric constraints $L_{plane}+L_n$, we provide a visual comparison of the reconstruction details in \cref{fig:meshing_geometric_constraints}. 
The reconstructed mesh without geometric constraints captures the coarse geometry and exhibits uneven and rough surfaces. 
In contrast, our method with geometric constraints recovers smoother and more accurate meshes. 
This indicates that incorporating geometric constraints guides Gaussians to better conform to the object surface, which ultimately improves the quality of reconstruction.

\begin{figure}[t]
    \centering
        \centerline{\includegraphics[width=0.9\textwidth]{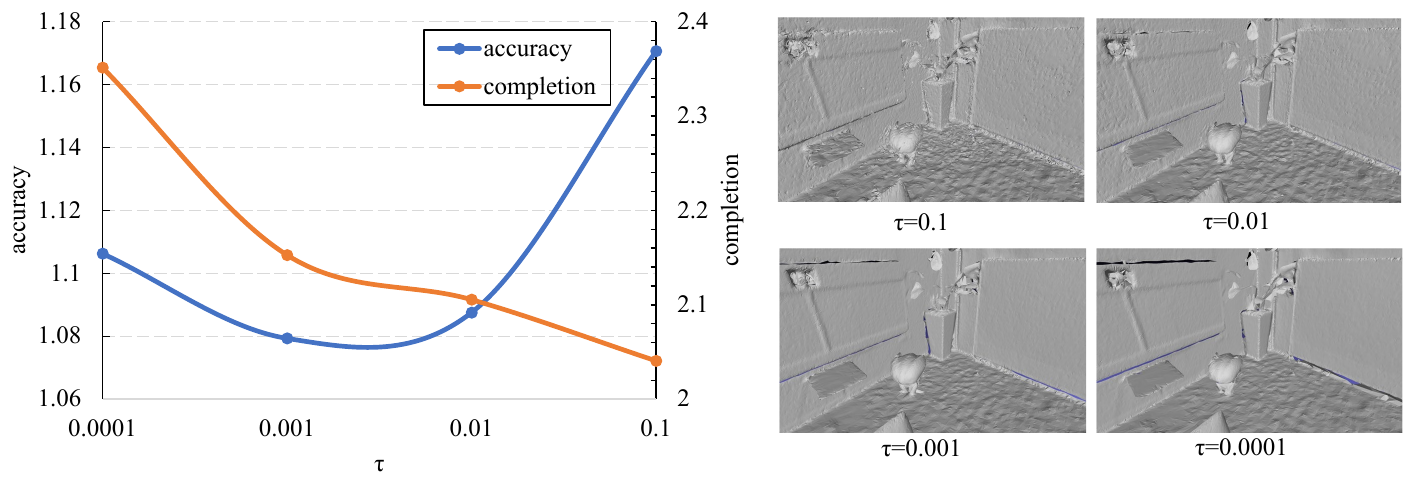}} 
    	\caption{Ablation Study on parameter $\tau$ to explore its impact on reconstruction quality.
     }
    	\label{fig:parameter_tau}
    
   \end{figure}

\noindent\textbf{Geometric Gaussian Set.} Since the quality of the geometric Gaussian set directly affects reconstruction, the parameter $\tau$ in the selection strategy significantly impacts mesh quality. To explore its influence on reconstruction, we conduct ablation on the Replica dataset with $\tau$ settings. We set $\tau$ ranging from 0.0001 to 0.1 across orders of magnitude. Accuracy and completion are used as reconstruction quality metrics, where lower values signify better performance. \cref{fig:parameter_tau} illustrates both quantitative and qualitative impacts of $\tau$. Increasing $\tau$ yields better completion but penalizes accuracy. Furthermore, the qualitative results demonstrate that improperly small values ($\tau=0.0001$) cause obvious missing geometry. Therefore, in our implementation, we adopt $\tau = 0.001$ to achieve an optimal balance between precision and completeness.

\section{Conclusion}
This paper proposed an incremental online scene reconstruction method using Gaussian triangulation. A dense geometric Gaussian representation was presented as an explicit and efficient foundation for both high-fidelity rendering and surface reconstruction. Building on this Gaussian representation, a novel direct meshing algorithm was proposed to incrementally reconstruct and update triangular meshes. To improve reconstruction accuracy, we introduced a plane-based pulling constraint that can dynamically align 3D Gaussians to local surface approximations. Furthermore, we achieved memory-efficient incremental reconstruction by progressively freezing fully optimized mesh regions. Experiments on public datasets demonstrate that our method incrementally recovers high-quality meshes online and simultaneously achieves high-fidelity rendering. One limitation of our method is its reliance on depth information and its inability to reconstruct unobserved regions. We hope to explore 3D reconstruction using only RGB information in future work.

\section*{Acknowledgements}
This work is supported by the National Natural Science Foundation of China under Grant No.62376244 and the National Key Research and Development Program of China (No. 2023YFF0905104). It is also supported by the Information Technology Center and State Key Lab of CAD\&CG, Zhejiang University.

\bibliographystyle{splncs04}
\bibliography{main}
\end{document}